\documentclass[10pt,twocolumn,letterpaper]{article}

\usepackage{conf}
\usepackage{times}
\usepackage{epsfig}
\usepackage{graphicx}
\usepackage{amsmath}
\usepackage{amssymb}

\usepackage{booktabs}
\usepackage{color}
\usepackage{multirow}
\usepackage{subfigure}
\usepackage[pagebackref=true,breaklinks=true,colorlinks,bookmarks=false]{hyperref}

\finalcopy

\newcommand{\green}[1]{\textcolor{green}{#1}}
\newcommand{\red}[1]{\textcolor{red}{#1}}

\def \m{\mathbf{m}}

\def \b{\mathbf{b}}

\def \x{\mathbf{x}}

\def \BA{\mathbf{A}}

\def \BX{\mathbf{X}}

\def \BI{\mathbf{I}}

\def \BD{\mathbf{D}}
\def \BE{\mathbf{E}}

\def \BG{\mathbf{G}}

\def \BW{\mathbf{W}}

\newcommand{\pare}[1]{{(#1)}}  

\begin{document}

\title{Diversified Multiscale Graph Learning with Graph Self-Correction}

\author{
\textbf{Yuzhao Chen,\textsuperscript{\rm 1,}\textsuperscript{\rm 2}\quad\quad
Yatao Bian,\textsuperscript{\rm 2}\quad\quad
Jiying Zhang,\textsuperscript{\rm 1,}\textsuperscript{\rm 2}\quad\quad
Xi Xiao,\textsuperscript{\rm 1}}\\
\textbf{
Tingyang Xu,\textsuperscript{\rm 2}\quad\quad
Yu Rong,\textsuperscript{\rm 2}\quad\quad
Junzhou Huang \textsuperscript{\rm 2,}\textsuperscript{\rm 3}}\\
\textsuperscript{\rm 1}Tsinghua Shenzhen International Graduate School, Tsinghua University, \\
\textsuperscript{\rm 2}Tencent AI Lab, Shenzhen, \\
\textsuperscript{\rm 3}University of Texas at Arlington, Arlington
}

\maketitle

\begin{abstract}
    Though the multiscale graph learning techniques have enabled advanced feature extraction frameworks, 
    the classic ensemble strategy may show inferior performance while encountering the high homogeneity of the learnt representation, which is caused by the nature of existing graph pooling methods.
To cope with this issue,
   we propose a diversified multiscale graph learning model equipped with two core ingredients:  a graph self-correction (GSC) mechanism to generate informative embedded graphs, and a diversity boosting regularizer (DBR) to achieve a comprehensive characterization of the input graph.
   The proposed GSC mechanism compensates the pooled graph with the lost information  during the graph pooling process by  feeding back  the estimated residual graph, which serves as a plug-in component for popular graph pooling methods.
Meanwhile, 
   pooling methods enhanced with the GSC procedure encourage the discrepancy of node embeddings, and thus it contributes to 
   the success of ensemble learning strategy.
   The proposed DBR instead enhances the ensemble diversity at the graph-level embeddings by leveraging the interaction among individual classifiers. 
   Extensive experiments on popular graph classification benchmarks show that the proposed GSC mechanism leads to significant improvements over state-of-the-art graph pooling methods. Moreover, the ensemble multiscale graph learning models achieve superior enhancement by combining both GSC and DBR.
  \vspace{-4mm}
\end{abstract}

\section{Introduction}

Graph Neural Networks (GNNs) have been developed rapidly for modeling graph-structured data and achieved remarkable progress in graph representation learning as well as 
downstream learning tasks, such as bioinformatics and social networks analysis~\cite{kipf2016semi,hamilton2017inductive,xu2018powerful,velivckovic2018graph}.
While many GNNs learn graph representation at a fixed scale, the multiscale graph learning has also attracted a surge of interests~\cite{chen2017harp,liang2018mile,deng2019graphzoom,gao2019graph,lee2019self,li2020graph}, for its capability of capturing both fine and coarse graph structures and features (representing local and global information, respectively).
Several advanced feature extraction frameworks have been proposed for multiscale graph learning.
Some attempts adopt the encoder-decoder pipeline to embed the input-graph and perform feature updating in the latent coarsest graph~\cite{chen2017harp,liang2018mile,deng2019graphzoom}. Other works utilize the pyramid architecture to extract multiscale graph features, and perform feature aggregation via skip-connection~\cite{gao2019graph,lee2019self}, or cross-layer summation fusion~\cite{li2020graph}.  

\begin{figure} [t]
  \centering
  \includegraphics[width=1.0\linewidth]{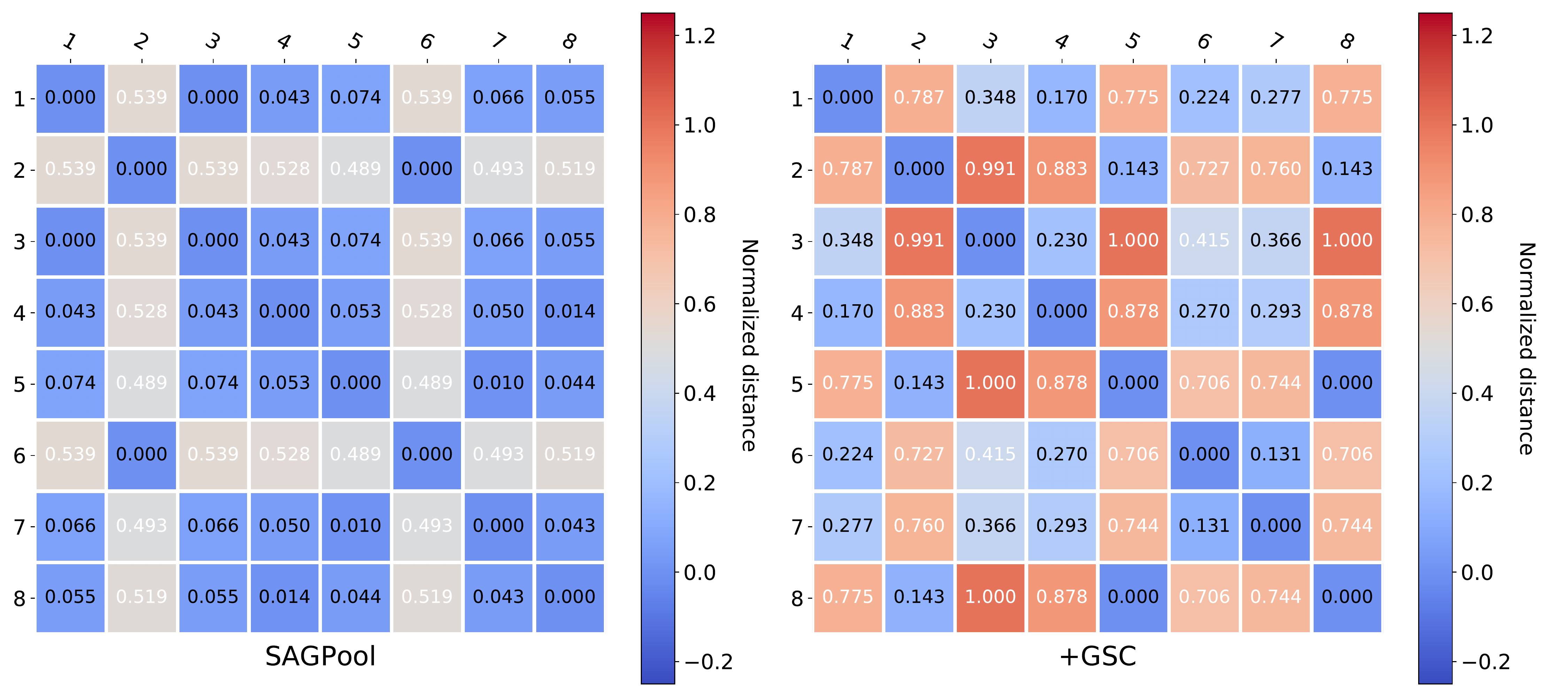}
  \caption{Distance heatmap of node features of the last graph pooling layers.
  Node features become highly homogenized in SAGPool~\cite{lee2019self} (left), while  graph self-correction well preserves node discrepancy (right) from the initial features (see Figure~\ref{fig:visual-embedding}).
 }
 \label{fig:visual-heatmap}
 \vspace{-3mm}
\end{figure}
    

Orthogonal to  various multiscale feature learning techniques, one straightforward and promising strategy for adequately leveraging the extracted graph embeddings is to construct ensembles of multiple individual classifiers in a stacking style, 
where each classifier receives a graph embedding at a fixed granularity, and predicted logits of all classifiers are then averaged to produce the final prediction. 
As a generic method, such ensemble models are expected to obtain more reliable predictions.    %
Meanwhile, the success of ensemble learning crucially depends on two properties of each individual classifier: high accuracy and high diversity~\cite{dietterich2000ensemble,ren2016ensemble}.
The first property is readily satisfied due to the high expressive power of deep neural networks~\cite{csaji2001approximation}, 
however, the second one may not hold in the scenario of multiscale graph learning. 

\begin{table}[t] 
\centering
\caption{
        The failing of the simple ensemble strategy (denoted with `+') on enhancing multiscale GNNs.
        Detail settings are stated in Section \ref{exp:exp-all}.
        More verifications are given in Appendix.
        }
\vspace{1mm}
\resizebox{0.9\linewidth}{!}{
    \begin{tabular}{c|ccc}
    \toprule
         Method                                           & PROTEINS                         & NCI1                         & NCI109             \\ 
    \midrule
        SAGPool                   &  73.26$_{\pm0.78}$ (-)                &  69.88$_{\pm0.82}$  (-)               &  70.07$_{\pm0.69}$ (-)      \\
        SAGPool+                    &  72.56$_{\pm0.76}$ (\green{$\downarrow$})   &  69.61$_{\pm0.62}$ (\green{$\downarrow$})  &  69.41$_{\pm0.66}$ (\green{$\downarrow$})      \\
    \midrule
        ASAP                    &  74.14$_{\pm0.33}$   (-)              &  74.27$_{\pm0.63}$   (-)                &  72.90$_{\pm0.59}$  (-)      \\
        ASAP+                  &  74.21$_{\pm0.69}$ (\red{$\uparrow$})       &  73.62$_{\pm0.56}$ (\green{$\downarrow$})       &  72.52$_{\pm0.45}$ (\green{$\downarrow$})        \\
    \bottomrule                     
    \end{tabular}                           
}
\label{tab:fail-ensemble}
\vspace{-3mm}
\end{table}

As indicated in recent researches~\cite{mesquita2020rethinking,bianchi2020spectral}, the indispensable graph pooling operation for establishing multiscale graph learning may have led to the high homogeneity of the node embeddings. 
Mesquita et al.~\cite{mesquita2020rethinking} observe that graph pooling via the learnable cluster assignment procedure~\cite{ying2018hierarchical, khasahmadi2020memory, bianchi2020spectral} exacerbate the homogeneity of node embeddings, and thus a comparable performance could be obtained by just replacing the complicated assignment matrix with some random matrix.
Bianchi et al.~\cite{bianchi2020spectral}  notice a similar issue in graph pooling methods based on the Top-$\textit{K}$ vertex selection mechanism~\cite{gao2019graph,lee2019self}. That is, these pooling layers would tend to select nodes that are highly-connected and share similar features.
Due to the homogeneity of the node embeddings,  the generated graph embeddings at various granularity tend to become homogeneous as well. 
Under this circumstance, 
these homogeneous graph embeddings lead to limited diversity among learned classifiers. 
Since  diversity is crucial for the success of ensemble methods,  
it is probable that  the ensemble strategy will fail in boosting the performance of  multiscale GNNs equipped with these pooling modules, or even become detrimental to them. 
This is verified by the experiments results given in Table \ref{tab:fail-ensemble}.

A common solution to improve the ensemble diversity is to build multiple classifiers that get trained with different sampled data in an independent manner. 
Nevertheless, it requires much more expensive computational cost
and
ignores the interaction among individual classifiers, which might also lead to shared feature representations among these members and would not be helpful to improve the performance~\cite{krogh1994neural}.
To address the issues discussed above, we design a diversified multiscale graph learning model that includes two main technical contributions: 
1) a graph self-correction mechanism to
generate informative pooled graphs, which also hampers the homogeneity of node embeddings and thus promotes the ensemble diversity at the $\textit{node-level embeddings}$, and
2) a diversity boosting regularizer to
directly model and optimize the ensemble diversity at the $\emph{graph-level embeddings}$.

\noindent\textbf{The graph self-correction mechanism.} 
It is intuitive to attribute the homogenization of node embeddings 
to the information loss caused by graph pooling operations.
Existing pooling methods mainly focus on how to construct the structure of the coarsened graph~\cite{ying2018hierarchical, khasahmadi2020memory, bianchi2020spectral,gao2019graph,lee2019self,huang2019attpool,lihierarchical2019}, either through a cluster assignment procedure or a Top-$\textit{K}$ vertex selection mechanism. 
However, few works have studied how to generate high-quality coarsened graphs with adequate semantic information.
We aim in this target by introducing 
Graph Self-Correcting (GSC) mechanism to compensate for the information loss during the graph pooling process.
As we will show, GSC works by moving the emphasis of the graph pooling problem from direct learning an embedded pooled graph to a two-stage  target: first, searching the optimal pooled graph structure, and second, self-correcting the semantic embeddings of it.
Notably, GSC not only enhances the embedding quality of the pooled graph by serving as a plug-and-play method, but also  relieves the homogeneity of node features (as shown in Figure~\ref{fig:visual-heatmap}), which further contributes to the success of the ensemble strategy on the multiscale graph learning model.

\noindent\textbf{The diversity boosting regularizer.} 

Another key ingredient of our contributions is to directly emphasize the diversity among the graph embeddings at different granularity.
It's also expected to provide a comprehensive characterization of the input graph from multiple diversified graph embeddings.
Specifically, we leverage the interaction among base learners and model the ensemble diversity on the graph embeddings of each scaled graph.
Under this definition, we propose a diversity boosting regularizer (DBR) to further diversify the multiscale graph learning model.
Note that previous work has devised ensemble diversity on the level of model predictions~\cite{islam2003constructive,pang2019improving,zhang2019nonlinear}, and we also make a discussion with them.

To validate the effectiveness of our methods, we conduct extensive experiments on popular graph classification benchmark datasets to perform a comparative study.
We show that our GSC leads to significant improvements over state-of-the-art graph pooling methods. Meanwhile, 
combing both GSC and DBR can consistently boost the ensemble multiscale graph learning model.

\section{Related Work}
\label{sec:related}

\subsection{Graph Pooling Modules}
For extracting multiple graphs at different scales, the graph pooling operation is essential to downsample the input full-graph. 
Since the graphs are often irregular data which lies on  the non-Euclidean domain, the locality of them is not well-defined. 
And hence, mature pooling techniques in Convolutional Neural Networks (CNNs) devised for processing regular data, such as image, can not be naturally extended to GNNs.
In some early attempts, researchers adopt some deterministic and heuristic methods to perform graph pooling.  Either applying  global pooling over all node embeddings~\cite{henaff2015deep} or coarsening the graph via traditional cluster assignment algorithms~\cite{bruna2013spectral,defferrard2016convolutional} are included in this aspect.
Further works focus on establishing the end-to-end hierarchical learning manner for embedding the graph
The first research line is distributed to  turning the problem of graph pooling  into  vertex  clustering through parameterizing the cluster assignment procedure so that the differentiable pooling operator is built. 
DiffPool~\cite{ying2018hierarchical} leverages the node features and graph topology to learn the soft assignment matrix for the first time. 
MinCUT~\cite{bianchi2020spectral} proposes a relaxed formulation of the spectral clustering and improve the  DiffPool to some extent.
GMN~\cite{khasahmadi2020memory} generates the assignment matrix using a sequence of memory layers.
Additionally, HaarPool~\cite{ma2019graph} and EigenPool~\cite{wang2020haar} are spectral-based pooling methods that use the spectral clustering to generate the coarsened graphs, but their time complexity costs are relatively higher.
Another technical line is based on the Top-\textit{K} vertex selection mechanism, which introduces a scorer model to calculate the importance score of each node,  and the nodes with \textit{K} highest scores are retained to construct the coarsened graph. 
gPool~\cite{gao2019graph} firstly uses the node features to train the scorer model, and SAGPool~\cite{lee2019self} improves upon it by further leveraging the graph topology for scoring nodes. 
AttPool~\cite{huang2019attpool} proposes a similar scoring model as SAGPool and a local-attention mechanism to prevent the sampled nodes from stucking in certain region.
ASAP~\cite{ranjan2020asap} proposes to update the nodes in the same cluster via a self-attention mechanism, and then scoring the nodes based on such enhanced node features. It also updates the adjacency matrix following DiffPool to maintain graph connectivity.
The Top-$K$ vertex selection based methods are usually more efficient since it avoids learning the dense assignment matrix.

\subsection{Feedback Networks}
Feedback networks decompose a one-step prediction procedure into multiple steps for establishing a mechanism of perceiving errors and making corrections. It has been used in many visual tasks~\cite{carreira2016human,haris2018deep,li2016iterative,lotter2016deep,shrivastava2016contextual,tu2009auto}. 
For the case of human pose estimation~\cite{carreira2016human}, the  feedback network iteratively estimates the position deviation and applies it back to the current estimation. DBPN~\cite{haris2018deep} proposes a projection module to involve iterative up- and down-sampling process in the super-resolution network.
To our knowledge, the notion of feedback procedures has not been applied to GNNs, especially the graph pooling operation.
This extension is technically non-trivial. While the offset or distortion produced in the prediction is clear for image signals, the irregular structures of graphs make  identifying what is missing or undesirable a challenge.
In this paper, we devise two schemes of graph self-correction and obtain considerable performance improvement, though we do not adopt an iterative process for the consideration of computational cost.

\section{Preliminaries}
\label{sec:preliminary}

Throughout this work, a graph is represented  as $\mathcal{G}(\mathcal{V}, \BA)$, 
where $\mathcal{V}$ is  vertex set that has $N$ nodes with $d_0$-dimension features of $\BX\in \mathbb{R}^{N\times d_0}$,
and $\BA\in \mathbb{R}^{N\times N}$ is the adjacency matrix. Note that we omit the edge features in this paper for simplicity.
The node degree matrix is given by $\BD=\mathrm{diag}(\BA \mathbf{1}_{N})$. 
The adjacency matrix with inserted self-loops is $\hat{\BA}=\BA+\BI$, and the corresponding degree matrix is denoted as $\hat{\BD}$.
Given a node $v$, its connected-neighbors are denoted as $\mathcal{N}_v$. For a matrix $\BX$, $\BX_{i,:}$ denotes its $i$-th row and $\BX_{:,j}$ denotes its $j$-th column.

\subsection{Multiscale Graph Neural Networks}   
The key operation of GNNs can be abstracted to a process that involves message passing  among the nodes in the graph. 
The message passing operation iteratively updates a node $v$'s hidden states, $\BX_{v,:}\in \mathbb{R}^{1\times d}$, by aggregating the hidden states of $v$'s neighboring nodes. 
In general, the message passing process involves several iterations, each one can be formulated as, 
\begin{align}\label{eq:message-passing}\notag
    & \m_v^{\pare{l}} = \mathrm{AGGR}^\pare{l}( \{ (\BX_{v,:}^\pare{l-1}, \BX_{u,:}^\pare{l-1}) \;|\;  u\in \mathcal{N}_v  \}  ),\\ 
    & \BX_{v,:}^\pare{l} = \sigma( \BW^\pare{l}\m_v^{\pare{l}} + \b^\pare{l}),
\end{align}
where  $\mathrm{AGGR}^\pare{l}(\cdot)$ is the aggregation operator, $\m_v^{\pare{l}}$ is the aggregated message,  $\sigma(\cdot)$ is some activation function
and $\BW$, $\b$ are the trainable parameters. 
Usually we set the initial hidden states $\BX^\pare{0}$ as node ego-features $\BX$.

Multiscale GNNs iteratively generate coarsened graphs to extract hierarchical representations, which are imperative for capturing both fine and coarse  structural and semantic information, by repeating hierarchical graph pooling operations.
Formally, assume there are $l_t$ number of message passing operations between $\mathrm{Pool}$ operation $P_{t-1}$ and $P_{t}$, i.e., the $t$-th $\mathrm{Pool}$ operation is acting on the $l_t$-th iterations of message passing on each scale's graph,
\begin{equation}\label{eq:hierarchical-pooling}
    \BX^{\pare{0, P_{t}}}, \BA^{\pare{0, P_{t}}} = \mathrm{Pool}^\pare{t}(\BX^{\pare{l_{t-1}, P_{t-1}}}, \BA^{\pare{l_{t-1}, P_{t-1}}}),
\end{equation}
where $t\in [1,T]$  represents the index number of the graph pooling layers.
We make a convention that $\BX^{\pare{l, P_{0}}}:=\BX^{\pare{l}}$ is the node feature matrix that has yet been pooled, and $\BX^{\pare{0, P_{t+1}}}:=\BX^{\pare{P_{t+1}}}$. 
Multiscale GNNs~\cite{deng2019graphzoom,lee2019self,li2020graph} utilize pooled graphs at multiple scales to perform feature fusion and generate final graph embedding,
\begin{align}\label{eq:readout}
    \x_G = \mathrm{READOUT}( f( \{ \BX^{\pare{P_{0}}}, ..., \BX^{\pare{P_{T}}}  \}) ),
\end{align}
where the fusion function $f(\cdot)$, such as skip-connection \cite{gao2019graph,lee2019self} or cross-scale summation \cite{li2020graph}, is used to aggregate multiscale graphs. The   following  $\mathrm{READOUT}$ is a permutation-invariant operator to get graph-level embeddings.
Lastly, a classifier such as the multi-layer perceptron $\mathrm{MLP}$ is used to predict the category logits
$\tilde{y} = \mathrm{MLP}(\x_G)$.

\section{Graph Self-Correction Mechanism}
\label{sec:method-1}
Existing graph pooling methods haven't yet focused on generating high-quality coarsened graphs with rich semantic information, 
and the information loss caused by pooling operations may have caused the homogenization of representations.
A core idea of Graph Self-Correction (GSC) mechanism  is to reduce such information loss through compensating information generated by the feedback procedures.
It contains three phases: graph pooling, compensated information calculation and information feedback.

\subsection{Phase 1: Graph pooling}
\label{sec:step1}
In the settings of GSC, the initial graph pooling layer concentrates on searching for the optimal structure of the coarsened graphs rather than directly learning both the topological and informative embeddings.
As a preliminary step of the approach, it can be implemented by any kind of existing pooling  methods~\cite{ying2018hierarchical,bianchi2020spectral,gao2019graph,lee2019self,huang2019attpool,ranjan2020asap}.
For the compactness of this paper, we only take the Top-$K$ vertex selection based pooling methods as the running example and we discuss the application of GSC on cluster assignment based methods in Appendix.

As a case of Eq.(\ref{eq:hierarchical-pooling}), SAGPool~\cite{lee2019self} adopts a $1$-layer GCN~\cite{kipf2016semi} as the node scorer, and uses the Top-$K$ selection strategy to select nodes retained in the coarsened graph,
\begin{align}
\label{eq:scorer}
    \mathbf{s}^{\pare{t}} &=\sigma(\hat{\BD}^{-\frac{1}{2}}\hat{\BA}^{\pare{l_t, P_t}}\hat{\BD}^{-\frac{1}{2}} \BX^{\pare{l_t, P_t}} \BW^{\pare{l_t}}), \\
\label{eq:topk}
    \text{Idx}^{\pare{t}} &= \mathrm{Top}\text{-}\mathrm{K}(\mathbf{s}^{\pare{t}}, K),
\end{align}
where the $\mathrm{Top}$-$\mathrm{K}$ function returns the $K$ indices of the selected nodes based on the ranking order of $\mathbf{s}^{\pare{t}}$,  which are then used to pool down the input-graph as:
\begin{align}
\label{eq:x-pool}
    \BX^{\pare{P_{t+1}}} &= \BX^{\pare{l_t, P_t}}_{\text{Idx},:} \odot \mathbf{s}_{\text{Idx}}, \\
\label{eq:a-pool}
    \BA^{\pare{P_{t+1}}} &= \BA^{\pare{l_t, P_t}}_{\text{Idx},\text{Idx}},
\end{align}
where $\odot$ is the element-wise product to apply node scores as attention weights for updating the pooled node features.

\subsection{Phase 2: Compensated information calculation}
As noticed in the pooling process given by SAGPool, the subsequent pooled node features discard all information preserved in those unselected nodes, which might hurt the exploitation of rich semantic features in original graphs. 
Inspired by the feedback networks~\cite{carreira2016human,haris2018deep,li2016iterative}, GSC introduces a residual estimation procedure to calculate compensated information that empowers the self-correction of the embedded coarsened graph in the feedback stage. 
Specifically, we propose two schemes to  calculate such 
residual 
signals: complement graph fusion and coarsened graph back-projection. 

\noindent\textbf{Complement graph fusion}
The graph composited by the unselected nodes after the graph pooling layer and their adjacency relations is defined as the complement of the pooled graph\footnote{In graph theory, the standard definition of the complement of a graph G is a graph H on the same vertices such that two distinct vertices of H are adjacent if and only if they are not adjacent in G.} here. Then the indices of nodes in the complement graph formulate:
\begin{equation}
    \text{CompIdx}=\{i(v)|v\in\mathcal{V} \; \text{and} \; i(v)\not\in\text{Idx}^{\pare{t}}\},
\end{equation}
where $i(\cdot)$ gives a unique index to a node. 
The complement denotes the information that has been lost during the first phase of GSC, and it can be adopted as the residual signal to be fused with the pooled graph.
The approach is to propagate node features from the  complement graph to the pooled graph leveraging an $\mathrm{UnPool}$ layer: 
\begin{equation}\label{eq:unpool-comp}
    \BE_{\text{Comp}}^{\pare{t}} = \text{UnPool}(\BX_{\text{Comp}}^{\pare{t}}, \mathbf{A}^{\pare{l_t, P_t}}).
\end{equation}
Here the $\mathrm{UnPool}$ denotes an unpooling process. It receives node features of the complement graph as the input feature and the original graph structure as adjacency relation, and interpolates compensated information to those selected nodes of the pooled graph.
Similar to \cite{gao2019graph}, it can be devised by initializing the input feature matrix $\BX_{\text{Comp}}^{\pare{t}}\in \mathbb{R}^{N\times d}$ as:
\begin{equation}\label{eq:feature-comp}
    \BX_{\text{Comp}}^{\pare{t}}\text{[}i(v),:\text{]} = 
        \begin{cases}
        \BX_{i(v),:}^{\pare{l_t,P_t}} & i(v)\in\text{CompIdx}\\
        \mathbf{0}                  & \text{otherwise},
        \end{cases}
\vspace{-1mm}
\end{equation}
here we denote the index selection operator in the brackets of the left-hand side for clarity, and then $\mathrm{UnPool}$ propagate it by the message passing process, which is implemented by a 1-layer graph convolution network (GCN)~\cite{kipf2016semi}.

\begin{figure*} [htbp]
  \centering
  \includegraphics[width=0.9\linewidth]{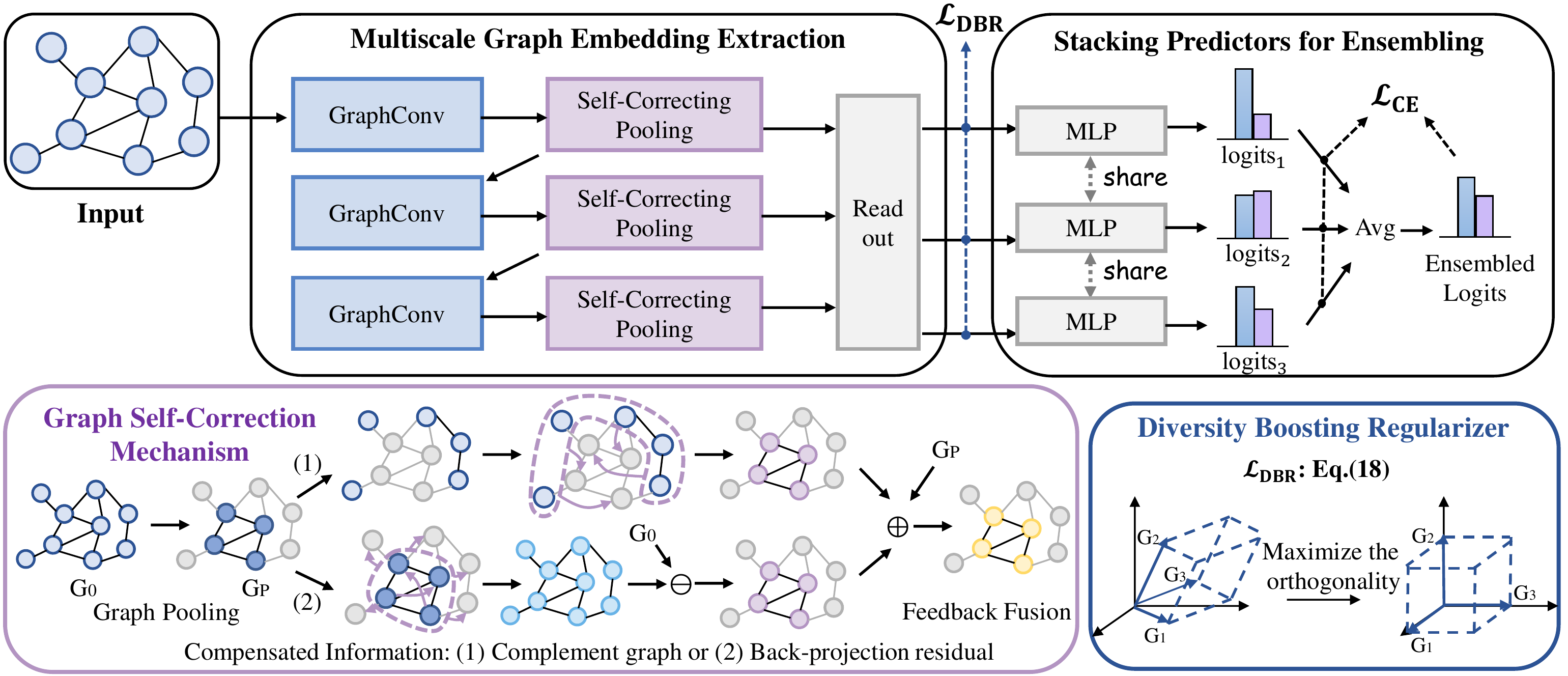}
  \caption{The ensembled multiscale graph learning framework (top), equipped with the proposed  graph self-correction mechanism (bottom left) and  diversity boosting regularizer (bottom right). 
The bottom-left module makes an illustration of the two modes of graph self-correction, and the bottom-right module provides a geometric interpretation of the regularizer.
 }
 \label{fig:framework} 
 \vspace{-2mm}
\end{figure*}
\noindent\textbf{Coarsened graph back-projection}
Another source of the compensated information is inspired by the
classic back-projection algorithm \cite{irani1991improving} used for iteratively refining the restored high-resolution images.
The intuition is that a pooled graph is ideal if it contains adequate information for reconstructing the original graph.
Rather than introducing additional knowledge to model the pattern of lost information, coarsened graph back-projection manages
to restore the original graph using the pooled node features themselves, and then calculates the reconstruction error to serve as the compensated signals.
The restoration process is given by:
\begin{equation}\label{eq:unpool-bp}
    \BX_{\text{Recon}}^{\pare{t}} = \text{UnPool}(\mathbf{X}_{\text{Coarse}}^{\pare{t}}, \mathbf{A}^{\pare{l_t, P_t}}),
\end{equation}
where the input feature matrix of $\mathbf{X}_{\text{Coarse}}^{\pare{t}}\in \mathbb{R}^{N\times d}$ is initialized by the pooled node features $\BX^{\pare{P_{t+1}}}\in \mathbb{R}^{|\text{Idx}|\times d}$ with padding zero vectors:
\begin{equation}\label{eq:feature-coarse}
\vspace{-1mm}
    \BX_{\text{Coarse}}^{\pare{t}}\text{[}i(v),:\text{]} = 
        \begin{cases}
        \BX_{i(v),:}^{\pare{t+1}} & i(v)\in\text{Idx}\\
        \mathbf{0}                  & \text{otherwise},
        \end{cases}
\end{equation}
After that, the residual graph is calculated by:
\begin{equation}\label{eq:residual-bp}
    \BE_{\text{Recon}}^{\pare{t}} = \mathbf{X}^{\pare{l_t, P_t}}-\BX_{\text{Recon}}^{\pare{t}}.
\end{equation}

\subsection{Phase 3: Information feedback}
GSC finally refines the pooled graph $\BX^{\pare{P_{t+1}}}$ with the residual embedded graph of $\BE^{\pare{l,P_t}}\in \mathbb{R}^{N\times d}$
as:
\begin{equation}\label{eq:sum-fusion}
    \Tilde{\BX}^{\pare{P_{t+1}}} =   \BX^{\pare{P_{t+1}}} + \BE_{\text{Idx},:}^{\pare{t}},
\vspace{-1mm}
\end{equation}
where $\BE_{\text{Idx},:}^{\pare{t}}\in \mathbb{R}^{|\text{Idx}|\times d}$ is tailored from the residual signal, either $\BE_{\text{Comp}}^{\pare{t}}\in \mathbb{R}^{N\times d}$ in Eq.(\ref{eq:unpool-comp}), or $\BE_{\text{Recon}}^{\pare{t}}\in \mathbb{R}^{N\times d}$ in Eq.(\ref{eq:residual-bp}), using the indices $\text{Idx}^{\pare{t}}$ given by Eq.(\ref{eq:topk}).

The graph self-correction mechanism not only helps the feedback of information vanished in the pooling process, but also reduces the homogeneity among nodes, which indicates that the self-correction learns to preserve the discrepancy of the original graph.
To verify this, we provide additional case studies and a deeper insight of the GSC mechanism in Section \ref{sec:how-geom-help-ensemble} for further discussion.

\section{Diversified Multiscale Graph Learning}
\label{sec:method-2}

The graph self-correction mechanism proposed in Section \ref{sec:method-1} generates informative multiscale embedded graphs, which also encourages the discrepancy of node-level embeddings and thus contributes to the ensemble learning strategy.
In this section, we first introduce the simple stacking-style ensemble multiscale graph learning model,
and then propose the diversity boosting regularizer acting on the readout graph embeddings,
which is working together with the GSC procedure to 
jointly enhance the diversity on node-level and graph-level representations.

\begin{table*}[htbp] 
\centering
\caption{Performance comparison of GSC mechanism with other pooling methods. 
        Methods with superscript $^*$ are the re-implemented version for a fair comparison.
        Methods denoted with `+' are the ensembled models.
        Average accuracy and standard deviation are reported.
        Best result on each comparison is bolded. 
        Best result of the non-ensembled models is underlined. 
        }
\vspace{1mm}
\resizebox{0.85\linewidth}{!}{
\begin{tabular}{c|cccccc}
\toprule
     Dataset                                     &   D\&D          & PROTEINS             & NCI1               & NCI109         &  FRANKENSTEIN   &  OGB-MOLHIV    \\ 
       \#Graphs                                &   1178        &   1113                &   4110            &   4127            &   4337        &       41127       \\
       Avg \#Nodes                             &   284.3       &   39.1               &   29.9           &   29.7           &   16.9           &       25.5        \\
\midrule
                    Set2Set~\cite{vinyals2015order}                      &  71.60 $\pm$ 0.87   &  72.16 $\pm$ 0.43   &  66.97 $\pm$ 0.74  &  61.04 $\pm$ 2.69  &  61.46 $\pm$ 0.47        &  --   \\
                    GlobalAttention~\cite{li2015gated}     &    71.38 $\pm$ 0.78    &  71.87 $\pm$ 0.60   &  69.00 $\pm$ 0.49   &  67.87 $\pm$ 0.40  &  61.31 $\pm$ 0.41             &  --   \\
                    SortPool~\cite{zhang2018end}           &    71.87 $\pm$ 0.96    &  73.91 $\pm$ 0.72   &  68.74 $\pm$ 1.07   &  68.59 $\pm$ 0.67  &  63.44 $\pm$ 0.65             &  --   \\
                 DiffPool~\cite{ying2018hierarchical} &    66.95 $\pm$ 2.41    &  68.20 $\pm$ 2.02   &  62.23 $\pm$ 1.90   &  61.98 $\pm$ 1.98  &  60.60 $\pm$ 1.62            &  --   \\
                          TopK~\cite{gao2019graph}             &    75.01 $\pm$ 0.86    &  71.10 $\pm$ 0.90   &  67.02 $\pm$ 2.25   &  66.12 $\pm$ 1.60  &  61.46 $\pm$ 0.84             &  --   \\
\midrule            
               SAGPool$^*$~\cite{lee2019self}   &     75.88 $\pm$ 0.72           &  73.26 $\pm$ 0.78                 &  69.88 $\pm$ 0.82                         &  70.07 $\pm$ 0.69                    &  60.68 $\pm$ 0.49                 &  73.16 $\pm$ 2.3  \\ 
            w/ GSC-B                  &     76.03 $\pm$ 0.68 (-)             &  74.27 $\pm$ 0.80 (-)                  &  71.91 $\pm$ 0.93    (-)                     &  71.69 $\pm$ 0.70      (-)               &  61.85 $\pm$ 0.79   (-)                &  73.53 $\pm$ 2.6  (-)  \\
               w/ GSC-B+             &    76.49 $\pm$ 0.96  (\red{$\uparrow$})             &  74.47 $\pm$ 0.72  (\red{$\uparrow$})                 &  $\textbf{73.10}$ $\pm$ 0.69  (\red{$\uparrow$})   &  72.13 $\pm$ 0.69    (\red{$\uparrow$})                       &    62.54 $\pm$ 0.47  (\red{$\uparrow$})                                &  72.73 $\pm$ 2.1  (\red{$\uparrow$})   \\
                w/ GSC-C   &     76.02 $\pm$ 0.67  (-)            &  74.36 $\pm$ 0.66   (-)              &  72.38 $\pm$ 0.69    (-)                     &  72.03 $\pm$ 0.70     (-)                    &  63.19 $\pm$ 0.62 (-) &  73.87 $\pm$ 1.6 (-) \\
               w/ GSC-C+     &    $\textbf{76.57}$ $\pm$ 1.04 (\red{$\uparrow$})   &  $\textbf{74.89}$ $\pm$ 0.70  (\red{$\uparrow$})  &  72.77 $\pm$ 0.73      (\red{$\uparrow$})             &   $\textbf{72.76}$ $\pm$ 0.57  (\red{$\uparrow$})  &  $\textbf{63.65}$ $\pm$ 0.54  (\red{$\uparrow$})       & $\textbf{74.01}$ $\pm$ 1.6  (\red{$\uparrow$})    \\
\midrule
               ASAP$^*$~\cite{ranjan2020asap}   &     76.77 $\pm$ 0.58              &  74.14 $\pm$ 0.33                  &  74.27 $\pm$ 0.63                            &  72.90 $\pm$ 0.59                     & 64.58 $\pm$ 0.23                      &  73.41 $\pm$ 2.2    \\
               w/ GSC-B                 &     76.80 $\pm$ 0.54  (-)            &  74.39 $\pm$ 0.71   (-)               &  74.52 $\pm$ 0.69  (-)                              &  $\underline{\text{73.64}}$ $\pm$ 0.49   (-)               & 64.34 $\pm$ 0.45  (-)                 &  $\underline{\text{74.99}}$ $\pm$ 1.7 (-)  \\
               w/ GSC-B+        &    $\textbf{77.46}$ $\pm$ 0.60  (\red{$\uparrow$})   &  74.86 $\pm$ 0.51  (\red{$\uparrow$})   &  $\textbf{74.93}$ $\pm$ 0.67  (\red{$\uparrow$})   &  $\textbf{74.32}$ $\pm$ 0.47  (\red{$\uparrow$})  &    65.01 $\pm$ 0.65   (\red{$\uparrow$})            &  $\textbf{75.31}$ $\pm$  1.5   (\red{$\uparrow$})   \\
               w/ GSC-C                  &     $\underline{\text{77.30}}$ $\pm$ 0.72 (-)  &  $\underline{\text{74.77}}$ $\pm$ 0.49  (-)                          &  $\underline{\text{74.83}}$ $\pm$ 0.41   (-)                              &  73.37 $\pm$ 0.58   (-)                   &  $\underline{\text{65.90}}$ $\pm$ 0.32  (-)        &  74.08 $\pm$ 1.8 (-)   \\
               w/ GSC-C+                &     77.35 $\pm$ 0.53  (\red{$\uparrow$})              &  $\textbf{75.15}$ $\pm$ 0.60  (\red{$\uparrow$})    & 74.84 $\pm$ 0.60   (\red{$\uparrow$})       &  73.57 $\pm$ 0.45    (\red{$\uparrow$})        &  $\textbf{66.17}$ $\pm$ 0.68   (\red{$\uparrow$})         & 74.76 $\pm$ 1.4 (\red{$\uparrow$})      \\
\bottomrule                     
\end{tabular}                           
}
\label{tab:main-comp}
\vspace{-3mm}
\end{table*}
\subsection{Model Architecture}
\label{sec:model-arch}
To fairly identify the contribution of each proposed ingredients, we design the multiscale graph learning framework following~\cite{gao2019graph,lee2019self,ranjan2020asap}.
In that case, $T$ and $L$ both equal to $3$ and $l_t$ is fixed to $1$ across all pooling layers.
The hierarchical pooled graphs at three scales are generated by three simple stacking modules of $\mathrm{GraphConv}$-$\mathrm{Pooling}$, then the graph embeddings are readout by the global pooling process separately.
We follow previous works to use a 1-layer GCN~\cite{kipf2016semi} followed by an activation function to work as $\mathrm{GraphConv}$.
$\mathrm{Pooling}$ consists of the baseline pooling module to be compared with and the proposed GSC mechanism. 
As a baseline backbone, the three graph embeddings are summed together before passing to an $\mathrm{MLP}$ to get the prediction logits.
For clarity, we provide the schemata of this baseline backbone in Appendix.

For performing the ensembled prediction, we modify the prediction module with three parameter-shared $\mathrm{MLP}$s, and each one receives a graph embedding from fixed granularity for predicting individual logits. 
Here the $\mathrm{MLP}$-sharing setting is for conducting rigorous controlled experiments.
After that, the ensembled logits are obtained by averaging three individual logits.
The model is depicted in Figure~\ref{fig:framework}. 
Noticed that the individual classifiers are a part of the multiscale learning architecture, thus they are trained on the same mini-batch of data but differ in the granularity of graphs to be readout as input. 
This feature enables us to utilize the interactions among these base learners and improves the training process, as stated in the  below section.

\subsection{Diversity Boosting Regularizers}

The collection of multiscale graph embeddings formulates $\BG = (\x_{G}^{\pare{P_1}}, \cdot\cdot\cdot, \x_{G}^{\pare{P_T}})^{\mathrm{T}}\in \mathbb{R}^{T\times d}$, where each element $\x_G^{\pare{P_t}}\in \mathbb{R}^{1\times d}$ is readout from $\{\BX_{v,:}^{\pare{P_t}}|v\in\mathcal{V}\}$ as a case of Eq.(\ref{eq:readout}).
Motivated by the theory of determinant point processes (DPPs)~\cite{kulesza2012determinantal}, 
we define the diversity of multiscale graph embeddings as:
\begin{equation}
\vspace{-1mm}
    \textit{DoG} = \mathrm{det}(\mathbf{\Sigma}),
\vspace{-1mm}
\end{equation}
where $\mathbf{\Sigma}$ is the similarity matrix of the multiscale graph embeddings. For example, it could be the gram matrix,  where each entry represents the inner-product similarity score of a pair of graph embeddings:
\begin{equation}
    \mathbf{\Sigma} = \tilde{\BG}\tilde{\BG}^{\mathrm{T}},
\vspace{-1mm}
\end{equation}
where $\tilde{\BG}$ is row-wise normalized from $\BG$, for guaranteeing the property of positive semi-definiteness on $\mathbf{\Sigma}$.
According to the matrix theory~\cite{zhang2011matrix}, the value of $\mathrm{det}(\BG\BG^{\mathrm{T}})$ equals to the squared volume of subspace spanned by graph embeddings $\{\x_G^{\pare{P_t}}|t\in\{1,...,T\}\}$, and hence $\textit{DoG}$ reaches the maximum value if and only if the graph embeddings are mutually orthogonal.

Noticed that the normalization of graph embeddings would reduce the variance of them, and lead the optimization problem to become trivial for regularizing the networks.
To address that, we introduce the gaussian kernel to parameterize the similarity matrix, which formulates:
\begin{equation}
\label{eq:gaussian}
    \mathbf{\Sigma}_{i,j}=\mathrm{exp}(-\gamma\cdot d^2(\BG_i, \BG_j)), \; i,j=1,\cdot\cdot\cdot,T,
\end{equation}
where $d(\cdot,\cdot)$ calculates the Euclidean distance and $\gamma$ is a hyper-parameter to control the flatness of the similarity matrix.
Under this definition, we propose the 
Diversity Boosting Regularizer (DBR)
to further diversify the multiscale graph embeddings,
\vspace{-1mm}
\begin{equation}
\label{eq:regularizer}
    L_{\text{DBR}}(\BG) = -\mathrm{log}(\textit{DoG}) + \mathrm{log}(\mathrm{det}(\mathbf{\Sigma}+\mathbf{I})),
\end{equation}
where the first term is the logarithm of embeddings diversity, 
the second term as normalization controls the magnitude  of the similarity matrix.
Although the training with DBR as regularizer involves the calculations of the matrix determinant and matrix inverse, it is still very efficient since the pooling number $T$ grows much slower with the scale of problem.
Lastly, for training the overall model, we combine the diversity regularizer $\alpha\cdot L_{\text{DBR}}$ ($\alpha$ as loss weight)  with the summed cross-entropy loss $\Sigma_T L_{\text{CE}}$ over all predicted logits as the training objective function.

\paragraph{Discussion with other diversity-based methods.}
We noticed that previous methods focusing on promoting the diversity of  ensemble models mainly consider the definition of ensemble diversity on the level of output predictions, including the classification error rates~\cite{islam2003constructive}, the normalized non-maximal classification scores~\cite{pang2019improving} and the predicted regression values~\cite{zhang2019nonlinear}. 
Notably, since many graph learning tasks are binary classification~\cite{morris2020tudataset,hu2020open}, encouraging either the error or non-maximal score of individual classifiers to be diverse certainly diminish the score of correct category, so that the loss on accuracy outweighs the benefit of enhanced diversity.
In the scenario of multiscale graph learning, we develop a new perspective of ensemble diversity, i.e. the  representation-level diversity, 
which not only wouldn't affect the classifier accuracy but promoting the representativeness of multiscale graph embeddings.

\vspace{-1mm}
\section{Experiments}
\label{exp:exp-all}

\subsection{Experiments Setup}
\paragraph{Dataset}
We consider six graph-level prediction datasets for conducting a comprehensive comparison. 
Five of them are part of the TU datasets~\cite{morris2020tudataset}: D\&D and PROTEINS are datasets containing proteins as graphs, NCI1 and NCI109 for classifying chemical compounds, FRANKENSTEIN possessing molecules as graphs,
and the recently proposed MOLHIV from Open Graph Benchmark (OGB)~\cite{hu2020open} and MoleculeNet~\cite{wu2018moleculenet} for identifying whether molecules as graphs inhibit HIV virus replication or not.
\vspace{-4mm}

\paragraph{Targets}
\label{sec:eval-baseline-protocol}
In the experiments, we aim at answering the following two questions:
Q1: Whether the Graph Self-Correction (GSC) mechanism can enhance graph pooling modules by serving as a plug-and-play method (Section \ref{exp:eva-GSC-main})?
Q2: Whether the two technical contributions (GSC and Diversity Boosting Regularizer (DBR)) can promote the success of ensemble learning (Section \ref{exp:eva-ensemble-main})?
\vspace{-1mm}

\subsection{Evaluation of Graph Self-Correction}
\label{exp:eva-GSC-main}
To answer the first question, 
we compare our GSC with previous graph learning methods, including hierarchical pooling based models of
DiffPool~\cite{ying2018hierarchical}, TopK~\cite{gao2019graph}, SAGPool~\cite{lee2019self},  ASAP~\cite{ranjan2020asap}, and global pooling based models of Set2Set~\cite{vinyals2015order}, GlobalAttention~\cite{li2015gated} and SortPool~\cite{zhang2018end}, 
under the non-ensemble architecture described in Section~\ref{sec:model-arch} (a minor difference is that global pooling methods only perform pooling after all GCN layers).
Among them, we select the two state-of-the-art pooling methods, SAGPool and ASAP as the evaluation baseline backbones to conduct in-depth comparisons.
Notably, SAGPool and ASAP both follow the same rigorous and fair evaluation protocol\footnote{Each experiment is run over 20 times of the 10-fold cross-validation under different random seeds, and especially avoids to perform  model selection on the testset but instead uses an additional validation set. 
The situations are somewhat different on some other graph classification networks~\cite{xu2018powerful,huang2019attpool,bianchi2020spectral,li2020graph},  which may have caused the inconsistent reported performance in the area.
Further discussion is deferred to Appendix.} for experiments on the TU dataset, 
and we strictly follow it. 
As for evaluation on MOLHIV, we use the default data split setting.
We use the same hyper-parameters search strategy (refer to Appendix) for all the baselines and our method. 

We report the results of the compared methods given in \cite{ranjan2020asap}, which are all evaluated in a comparable manner.
And we re-implement the selected baselines of SAGPool and ASAP to eliminate the differences caused by different experimental frameworks or other minor details for a fair comparison, which have obtained performances on par with or better than theirs. 
We denote GSC that calculates the residual signal $\BE_{\text{Comp}}$ from complement graph  as `GSC-C', and the one $\BE_{\text{Recon}}$ from graph back projection   as `GSC-B'.
The overall results are summarized in Table \ref{tab:main-comp}.

One can observe that the GSC mechanism achieves consistent and considerable performance improvement for all the cases.
It improves around $1.5\%$ accuracy and $1.0\%$ accuracy over six benchmarks on the baseline of SAGPool and ASAP, respectively.
Generally, the complement graph fusion (`GSC-C') performs better than the back projection (`GSC-B') on the SAGPool method, because the complement graph provides the pooled graph with richer information of the semantic representations of unselected nodes.
Instead, in the cases of NCI109 and MOLHIV on ASAP, GSC based on back projection (`GSC-B') achieves superior performance,
which might indicate that the reconstruction residuals have well refined the updating pooled node features generated by the graph pooling process.
We also employ a statistical test  to provide a more convincing indication of the gain brought by GSC in Appendix.

\begin{figure}[htbp]
\begin{center}
\includegraphics[width=1.0\linewidth]{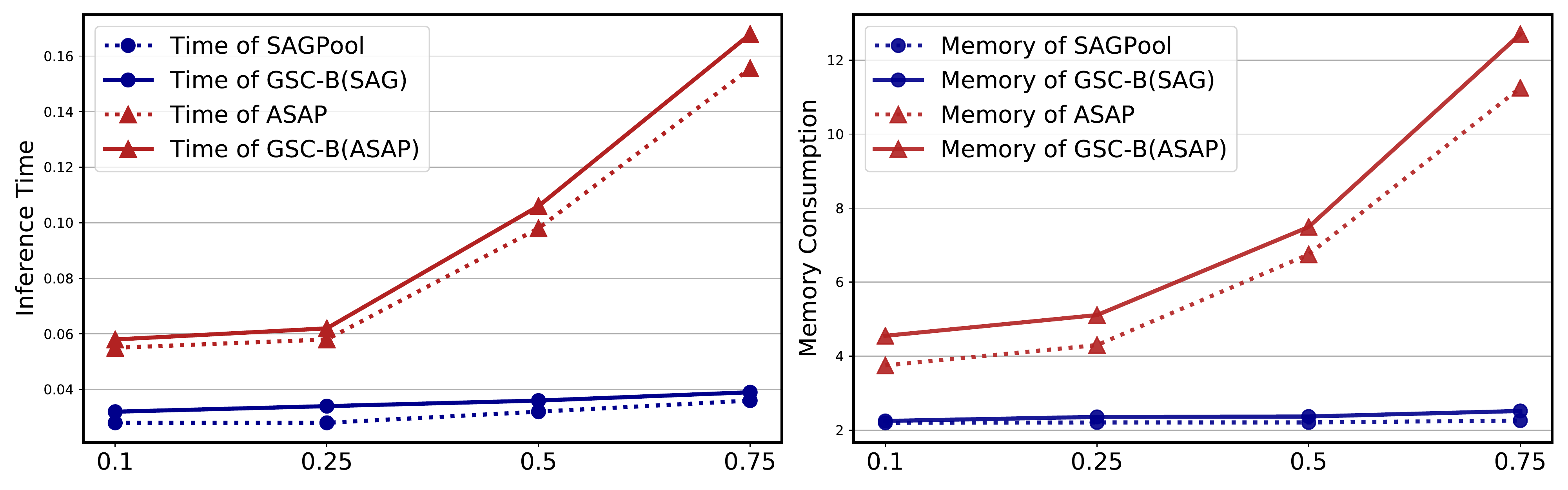}
\end{center}
\vspace{-4mm}
\caption{Time (left), Memory (right) curve w.r.t sampling ratio.}
\label{fig:long}
\vspace{-2mm}
\end{figure}

\begin{table}[htbp] 
\centering
\caption{Comparison for self-correction and post-enhancement. The number in parentheses refers to the relative improvement.}
\vspace{1mm}
\resizebox{0.88\linewidth}{!}{
\begin{tabular}{c|ccccc}
\toprule
     Method                                           & PROTEINS             & NCI1               & NCI109             \\ 
\midrule
    SAGPool                         &  73.26 (-)                &  69.88 (-)               &  70.07 (-)      \\
    w/ PGPE                    &  73.35 (+0.11)        &  70.32 (+0.44)        &  70.30  (+0.23)    \\
    w/ GSC                  &  $74.36$ ($\textbf{+1.10}$)          &  $72.38$ ($\textbf{+2.50}$)        &  $72.03$  ($\textbf{+1.96}$)     \\
\bottomrule                     
\end{tabular}                           
}
\label{tab:comp-post}
\vspace{-3mm}
\end{table}

\vspace{-5mm}
\subsubsection{Complexity Analysis}
The graph self-correction needs to additionally add the steps of compensated information calculation and feedback.
The time complexity of phase 1 depends on the based pooling module. It takes $O(Nd^2)+O(\vert\mathcal{E}\vert d)$ to derive the compensated graph in phase 2, and $O(Nd)$ to perform the fusion in phase 3.
We further analysis the trade-off between complexity and accuracy  through the following three aspects. 

First,
we compare the inference time and the training memory consumption of the methods. 
As shown in Figure \ref{fig:long},
GSC only increases marginal computational overhead over the baselines.
Second, we compare the efficiency of GSC and the most competitive module of ASAP \cite{ranjan2020asap}. Using SAGPool \cite{lee2019self} as a baseline, and a metric defined as the average accuracy improvement $\Delta_{\text{Acc}}$ divided by increased time cost $\Delta_{\text{Time}}$, GSC achieves a score of $3.75$ while ASAP  only obtains $0.33$. 
We provide detailed results in Appendix.
Third, we evaluate the contribution of the increased model capacity.
We establish another baseline module named as Pooled Graph Post-Enhancement (PGPE), which  leverages a single GCN layer to update the coarsened graph after each hierarchical pooling layer. 
In this case, the GNNs equipped with PGPE and GSC possess the same number of parameters, whose results are summarized in Table~\ref{tab:comp-post}.
Clearly, the post-enhancement strategy (`w/ PGPE') fails in improving the embedding quality  and obtains slight performance gain, while the self-correction during pooling process (`w/ GSC') is much more effective and achieves over 1.5\% accuracy improvement on average.

\begin{figure*} [htbp]
  \centering
  \includegraphics[width=0.8\linewidth]{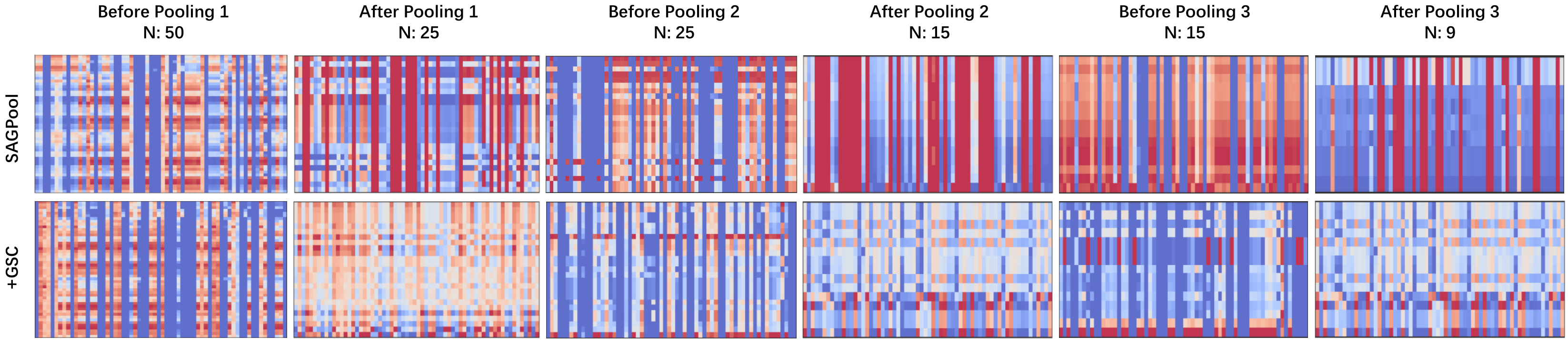}
  \caption{Visualization of node embeddings of a graph before and after pooling layers on PROTEINS. 
  Visualization maps are stretched to be vertically aligned, and a node is represented by a row.
  The number of nodes for each embedding matrix is given above the figure.
  On the SAGPool baseline, the high homogeneity of node embeddings is observed (the same as findings in \cite{mesquita2020rethinking}), meaning the graph pooling process exacerbate the information loss.
  The GSC instead preserves more diversified representation patterns from the original features. 
 }
 \label{fig:visual-embedding}
 \vspace{-2mm}
\end{figure*}
\subsection{Evaluation of Ensemble Muitiscale Learning}
\label{exp:eva-ensemble-main}
Next we answer to the second question by performing evaluations on the ensemble framework depicted in Figure \ref{fig:framework} and validating the effect of GSC and DBR.  
\vspace{-3mm}
\subsubsection{How the GSC Helps the Ensemble Learning}
\label{sec:how-geom-help-ensemble}
\textbf{Quantitative result.}
We particularly compare the ensemble performance 
between with and without graph self-correction mechanism on the pooling process.
The results are separately given in Table \ref{tab:main-comp} (methods denoted with `+') and Table \ref{tab:fail-ensemble}. 
Table \ref{tab:fail-ensemble} illustrates the failure of ensemble strategy on boosting multiscale GNNs built on baseline pooling modules. 
In contrast, for the multiscale GNNs equipped with GSC procedure (`GSC-C' or `GSC-B'), 
the ensemble strategy (`GSC-C+' or `GSC-B+') successfully achieves over $0.5\%$ average accuracy improvement 
on the even stronger baselines. 
More experimental results under different model settings are given in Appendix.

\noindent\textbf{Qualitative analysis.}
Intuitively, the failure of classic ensemble strategy on multiscale GNNs is natural if the node representations possess similar patterns, since the stacking-style ensemble learning requires diversified information.
We conduct an exploring analysis to validate this intuition.
Figure \ref{fig:visual-embedding} shows the phenomenons that are in line with our expectations: nodes become severely homogenized after the second graph pooling process with SAGPool~\cite{lee2019self}, while the GSC mechanism helps relieve this issue. 
The visualization analysis has provided a deeper insight into the beneficial effect of graph self-correction on the ensemble multiscale learning, that is,  
promoting the preserving of  nodes discrepancy in the coarse graph via restoring the fine graph, which further helps to increase the diversity of input information of the ensemble model.
\vspace{-4mm}

\begin{figure} [thbp]
  \centering
  \includegraphics[width=0.9\linewidth]{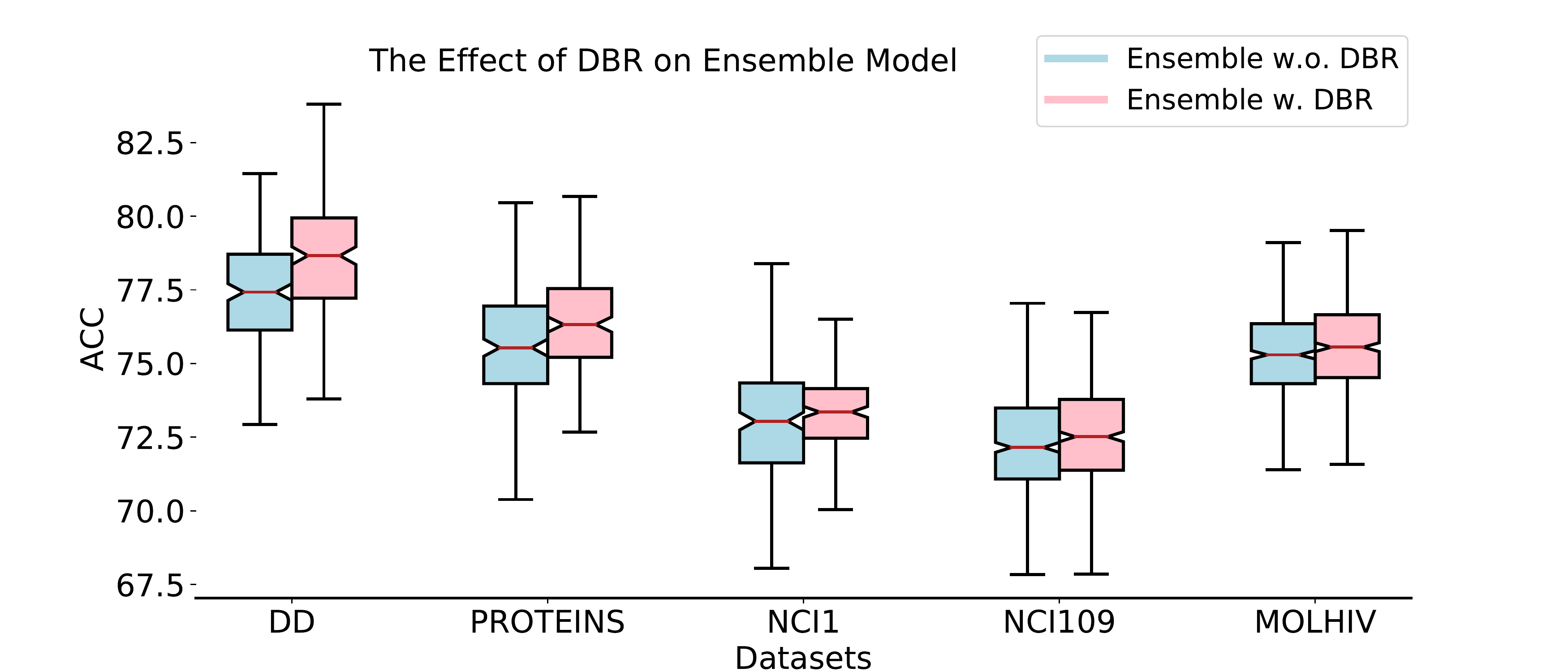}
  \caption{Comparison between ensemble models training with (pink) and without (lightblue)  the diversity boosting regularizer. Hyper-parameter settings are deferred to Appendix.
  The box plot shows the mean accuracy and the standard deviation. 
 }
 \label{fig:effect_dbr}
 \vspace{-3mm}
\end{figure}
\vspace{-1mm}
\subsubsection{How the DBR Helps the Ensemble Learning}
\label{sec:exp-dbr}
We conduct a comparison study on the ensemble multiscale GNNs equipped with GSC, under the difference between training with and without the proposed DBR.
The results are displayed in Figure~\ref{fig:effect_dbr}, which verify that DBR can jointly  improve the ensemble performance with the GSC mechanism. 
One can see that DBR achieves smaller standard deviation under the 10-fold cross validation, which means the diversified graph embeddings indeed provide a  more comprehensive characterization of the input graph, and thus
improve the training stability.
The effectiveness of DBR and GSC on enhancing the representation diversity provides a general and effective solution to build the ensemble-based multiscale graph classification networks.
\vspace{-1mm}

\begin{table}[thbp] 
\centering
\caption{The effect of DBR for exceeding SOTAs on MOLHIV.}
\vspace{1mm}
\resizebox{0.8\linewidth}{!}{
\begin{tabular}{c|ccccc}
\toprule
     Method                     & Baseline             & + Ensemble               & + DBR            \\ 
\midrule
    GCN                     &  $76.16_{\pm0.96}$ &  $77.54_{\pm0.83}$   &  $\textbf{78.06}_{\pm1.13}$      \\
    DeeperGCN               &  $77.50_{\pm1.71}$ &  $78.63_{\pm0.86}$    &  $\textbf{79.36}_{\pm1.43}$      \\
\bottomrule                     
\end{tabular}                           
}
\label{tab:SOTA-OGB}
\vspace{-3mm}
\end{table}
\vspace{1mm}
\noindent\textbf{Performance on Deeper GNNs.}
We provide additional experiments to apply  DBR for exceeding state-of-the-art (SOTA) GNNs on MOLHIV, which are usually deeper networks and adopt the global pooling layer for generating graph-level embeddings, since the dataset have graphs with fewer nodes.
We select the GCN~\cite{kipf2016semi} and DeeperGCN~\cite{li2020deepergcn} as evaluation baselines, and construct the ensemble-style model on them (details are deferred to Appendix. 
The results are shown in Table \ref{tab:SOTA-OGB}. 
It's clear to observe performance improvements by applying the diversity boosting regularizer on these SOTA models, which again verifies its effectiveness and generalization ability.

\section{Conclusion}
Compared to various multiscale graph feature extraction frameworks, we provide an orthogonal perspective to establish a practical ensemble model. The proposed graph self-correction not only leads to significant improvements over existing graph pooling methods by serving as a plug-in component, but also contributes to promoting the ensemble diversity on node-level embeddings. Working together with the new diversity boosting regularizer that enhances diversity on graph-level embeddings, they jointly lead the ensemble model to achieve superior performances.

{\small
\bibliographystyle{ieee_fullname}
\bibliography{egbib}
}

\end{document}